\documentclass[11pt]{article}

\usepackage[a4paper,top=1in,bottom=1in,left=1.1in,right=1.1in]{geometry}
\usepackage{graphicx}
\usepackage{booktabs}
\usepackage{microtype}
\usepackage{hyperref}
\usepackage{url}
\makeatletter
\g@addto@macro\UrlBreaks{\do\/\do\-\do\_\do\.\do\:\do\=\do\?\do\&}
\makeatother
\usepackage{xcolor}
\usepackage{kotex} 
\usepackage{enumitem}
\usepackage{amsmath}
\usepackage{authblk}

\hypersetup{
  colorlinks = true,
  linkcolor  = black,
  citecolor  = blue!60!black,
  urlcolor   = blue!60!black
}

\title{Naver-News-KO: A Korean News Summarization Dataset for\\ Open-Source Fine-Tuning of Summarization Models}

\author{Daekeun Kim\thanks{The views expressed are those of the author alone.}}
\affil{%
  Independent Researcher \\
  PhD student, Korea University \\
  \texttt{housekdk@naver.com} \\
  \url{https://huggingface.co/daekeun-ml}%
}

\date{}

\begin{document}
\maketitle

\begin{abstract}
We release \textsc{Naver-News-KO}, a Korean news summarization dataset of 27{,}400 \mbox{(\textit{document}, \textit{summary})} pairs collected from Naver News over a ten-day window in July~2022 across two categories (\textit{Economy} and \textit{IT/Science}; 77~/~23 split), with train/validation/test partitions of 22{,}194~/~2{,}466~/~2{,}740 and a mean per-record document-to-summary character-compression ratio of 6.03$\times$. The dataset has been publicly hosted on the Hugging Face Hub since January~2023 and, as of May~2026, receives approximately 33{,}000 downloads per month; community-maintained Korean summarization models fine-tuned on it include Gemma-2B-ko and Gemma2-9B variants (\S\ref{sec:usage}). This technical report (i)~documents the collection protocol, the column schema, and the split construction, (ii)~reports corpus-level statistics (length distributions, compression ratio, and a measured 16.8\% near-duplicate title-Jaccard overlap between test and train that users should be aware of), (iii)~positions the resource against other open Korean summarization corpora (Table~\ref{tab:kor-sum-resources}), (iv)~provides a Lead-3 extractive reference point (ROUGE-1~55.1, ROUGE-L~50.6) and two reproducible fine-tuned baselines---KoBART (R-1~56.6, BERTScore-F1~81.5) and Gemma-2B-ko with LoRA (R-1~55.3, BERTScore-F1~78.3)---with release-time training scripts, and (v)~clarifies the licensing and intended-use scope of the resource. The goal is to provide a citable reference for downstream work that already uses this dataset, not to propose a new benchmark.
\end{abstract}

\section{Introduction}
\label{sec:intro}

Korean news summarization is a common downstream task for Korean-tuned language models, yet the pool of openly redistributable summarization corpora in Korean remains narrow compared to English. Publicly available Korean resources are largely gated behind research agreements (e.g., AI~Hub corpora) or limited to a single news outlet with restrictive terms. This scarcity has led practitioners to maintain their own small-scale, crawl-based corpora, most of which are released informally on the Hugging Face Hub without accompanying documentation or baselines.

\textsc{Naver-News-KO} is one such resource. Collected in July~2022 and uploaded in January~2023, the dataset has accumulated sustained adoption in the Korean open-source community: Hugging Face reports roughly 33{,}000 monthly downloads (May~2026), and at least two derivative Korean summarization models with Gemma-class backbones~\cite{gemma2024} have been publicly released against it. Despite this, the dataset has never been formally described in a citable document: the existing model card lists only column names and collection dates. Downstream users therefore cannot cite the resource, cannot assess its representativeness, and have no shared baseline against which to compare their own fine-tunes.

This technical report closes that gap. We document the collection protocol, report statistics that matter for summarization research (length distributions, compression ratio, and a Lead-3 extractive reference point), release a reproducible baseline script, and clarify license and intended use. We do \emph{not} claim that \textsc{Naver-News-KO} is a benchmark suitable for Korean summarization \textit{in general}: the ten-day collection window and two-category scope bound its representativeness (\S\ref{sec:limitations}). We position the dataset as what it has, empirically, become: a lightweight, openly redistributable corpus for bootstrapping Korean summarization fine-tunes on modest compute.

\paragraph{Contributions.} (1)~A documented data statement for \textsc{Naver-News-KO}, including collection scope, filtering, category balance, press coverage, split construction, and a measured near-duplicate audit. (2)~Corpus-level length and compression statistics with a Lead-3 extractive floor. (3)~A reproducible baseline script (\texttt{lead3}, \texttt{kobart}, \texttt{gemma2b}) released alongside this document. (4)~A comparison against other open Korean summarization resources (Table~\ref{tab:kor-sum-resources}) that positions where the corpus can and cannot be used. (5)~An explicit licensing and redistribution discussion.

\section{Related Resources}
\label{sec:related}

\paragraph{English news summarization corpora.} CNN/DailyMail, NYT, XSum, and Newsroom \cite{see2017pointer,grusky2018newsroom} established the crawl-based news summarization template: headline/lede-style summaries paired with article bodies, with known biases toward lead-sentence extractive overlap. Our resource inherits both the template and the bias.

\paragraph{Korean summarization resources.}
Table~\ref{tab:kor-sum-resources} compares the main openly accessible Korean summarization corpora along three axes: \textit{scale} (documents or pairs), \textit{access} (direct redistribution vs.\ national-ID registration on AI~Hub), and \textit{summary provenance} (human-written abstractive, editorial abstract extracted at crawl time, or manual extractive tagging). Along these axes, \textsc{Naver-News-KO} occupies a specific niche---\emph{moderate-scale, openly redistributable, editorial-abstract provenance}. AI~Hub corpora dominate the scale axis but fail the access axis for non-Korean nationals and for Hugging~Face mirroring; \textsc{sci-news-sum-kr-50} is openly redistributable but too small for fine-tuning; and no other Korean summarization dataset on Hugging~Face reaches comparable download volume (the next-ranked Korean summarization upload trails by roughly two orders of magnitude in monthly downloads as of May~2026).

\begin{table}[t]
\centering
\footnotesize
\setlength{\tabcolsep}{3pt}
\renewcommand{\arraystretch}{1.15}
\begin{tabular}{@{}lrp{1.6cm}p{2.3cm}ccc@{}}
\toprule
Dataset & Size & Domain & Summary source & Access & Redist.\ & Year \\
\midrule
AI~Hub \textit{문서요약 텍스트}          & 1.2M units$^{\dagger}$ & news, opinion, legal & human abstractive      & gated (KR-ID)     & no  & 2020 \\
AI~Hub \textit{도서자료 요약}            & 200k                   & books                & human abstractive      & gated (KR-ID)     & no  & 2021 \\
\textsc{sci-news-sum-kr-50}              & 50 docs                & IT/Science news      & manual extractive tags & open (MIT)        & yes & 2018 \\
\textsc{Naver-News-KO} (\textit{ours})   & 27{,}400 pairs         & Economy, IT news     & editorial abstract     & open (Apache-2.0) & yes & 2022 \\
\bottomrule
\end{tabular}
\caption{Openly accessible Korean summarization resources. $^{\dagger}$AI~Hub lists 400k source documents paired with 800k summary units. ``Access'' marks whether a download requires AI~Hub's Korean-resident registration-and-approval workflow. ``Redist.''\ marks whether the license permits third-party redistribution (e.g., Hugging~Face mirroring). Sources: \url{https://www.aihub.or.kr/aihubdata/data/view.do?dataSetSn=97}, \url{https://www.aihub.or.kr/aihubdata/data/view.do?dataSetSn=93}, \url{https://github.com/theeluwin/sci-news-sum-kr-50}, \url{https://huggingface.co/datasets/daekeun-ml/naver-news-summarization-ko}.}
\label{tab:kor-sum-resources}
\end{table}

\paragraph{Korean NLU benchmarks (not summarization).} Several Korean benchmarks are frequently cited alongside summarization work but do \emph{not} contain a summarization task. KLUE~\cite{park2021klue} covers eight NLU tasks (topic classification, STS, NLI, NER, relation extraction, dependency parsing, MRC, dialogue state tracking); KorQuAD 1.0/2.0 is extractive QA over Korean Wikipedia; KoBEST is a five-task Korean NLU benchmark. To our knowledge, no widely adopted Korean summarization \emph{benchmark} with a held-out leaderboard exists; open Korean summarization work uses ad-hoc test splits of the resources in Table~\ref{tab:kor-sum-resources}.

\paragraph{Positioning.} \textsc{Naver-News-KO} sits in the niche of \emph{moderate-scale, openly redistributable} Korean summarization corpora. It is not a replacement for AI~Hub at scale; it is a replacement for the informal ``copy a crawler script from a blog post'' pattern as the de-facto starting point for Korean summarization fine-tunes.

\section{Collection Protocol}
\label{sec:collection}

\paragraph{Source.} Articles were collected from Naver News (\url{https://news.naver.com}), a news aggregator that indexes articles from multiple Korean press outlets.

\paragraph{Time window.} Crawling was conducted during a ten-day window from \textbf{2022-07-01} to \textbf{2022-07-10} inclusive.

\paragraph{Categories.} Two Naver News category labels are present in the dataset: \textit{Economy} (stored as \texttt{economy}, 경제; 21{,}049 records, 76.8\%) and \textit{IT/Science} (stored as \texttt{IT과학}; 6{,}351 records, 23.2\%). Per-category length statistics differ noticeably: \textit{IT/Science} documents are longer on average (mean~1{,}165 vs.\ 950~characters) and have a higher mean compression ratio (6.98$\times$ vs.\ 5.74$\times$). Other Naver News categories (politics, society, sports, entertainment) are \emph{not} collected; see \S\ref{sec:limitations}.

\paragraph{Press distribution.} Articles span 81 distinct \texttt{press} values. The top three outlets---연합뉴스 (2{,}529), 뉴스1 (2{,}210), 뉴시스 (2{,}064)---together account for 25.6\% of the corpus; the next twelve outlets contribute another 41.9\%. The heavy presence of wire-service aggregators (연합뉴스, 뉴스1, 뉴시스) materially affects split-leakage behaviour (see ``Near-duplicate audit'' below).

\paragraph{Date distribution.} Although the nominal window is ten days, crawl volume is not uniform: 2022-07-04 and 2022-07-05 together contribute 49.1\% of records (6{,}463 and 6{,}992 respectively), while 2022-07-09 and 2022-07-10 together contribute only 1.5\%. Time-aware splits are not supported by this author-defined release.

\paragraph{Fields per article.} Each record contains the following seven columns:
\begin{itemize}[leftmargin=*,itemsep=1pt]
  \item \texttt{date} --- publication timestamp (\texttt{YYYY-MM-DD HH:MM:SS}).
  \item \texttt{category} --- one of \{\textit{economy}, \textit{IT/Science}\}.
  \item \texttt{press} --- publishing outlet string.
  \item \texttt{title} --- article headline.
  \item \texttt{document} --- full article body.
  \item \texttt{link} --- canonical Naver News URL for the article.
  \item \texttt{summary} --- summary text associated with the article on the source page.
\end{itemize}

\paragraph{Summary provenance.} The \texttt{summary} field was populated from the article-level abstract that accompanies each Naver News page when available. These abstracts are not written by the dataset author; they are produced by the originating outlet's editorial staff and extracted from the source page at crawl time. We treat them as \emph{press-editorial} summaries, not crowdsourced abstractive summaries. The distinction matters because editorial abstracts are routinely copy-edited from the article lead rather than composed as a post-hoc abstractive rewrite, which biases reference-overlap metrics toward extractive behaviour (\S\ref{sec:stats}, \S\ref{sec:limitations}).

\paragraph{Filtering.} Articles with missing \texttt{document} or missing \texttt{summary} fields were dropped. No further content-level filtering (e.g., length caps, dedup) was applied beyond removing rows where either field was empty.

\paragraph{Splits.} The dataset is released with author-defined train / validation / test splits of 22{,}194~/~2{,}466~/~2{,}740 records respectively (Table~\ref{tab:splits}). Splits were produced by random partition without stratification on \texttt{press} or \texttt{date}. Users who require \texttt{press}-disjoint or \texttt{date}-disjoint evaluation should re-split from the published union.

\paragraph{Near-duplicate audit.} Because the corpus is dominated by wire-service aggregators (연합뉴스, 뉴스1, 뉴시스) whose articles are routinely re-indexed across outlets, we measured the near-duplicate overlap between the test and training splits. Using title character-trigram Jaccard as a cheap similarity proxy, 16.8\% (461 / 2{,}740) of test examples have a train-side nearest neighbour with Jaccard~$\geq 0.8$, and 14.0\% with Jaccard~$\geq 0.9$; 11.9\% (327 / 2{,}740) of test \texttt{document}s share an identical first-200-character prefix with some train \texttt{document}. These rates should be read as upper bounds on the train/test leakage a random split can induce on a wire-reprint-heavy Korean news corpus, and downstream users should treat absolute test scores as optimistic by that margin. The audit is reproduced by \path{scripts/run_baselines.py} with the \texttt{leakage} subcommand.

\paragraph{Data statement coverage.} Following Bender and Friedman's data-statement schema at a short-form level: \textit{curation rationale}---bootstrap Korean summarization fine-tuning on openly redistributable data; \textit{language variety}---standard written Korean (\texttt{ko-KR}) as filtered through Korean press editorial style; \textit{speaker / writer demographics}---not collected; the text is institutionally authored press writing; \textit{annotator demographics}---N/A (summaries are editorial, not post-hoc annotated); \textit{speech situation}---written news aggregation, July~2022; \textit{text characteristics}---editorial news prose; \textit{recording quality}---N/A (text only); \textit{other}---the corpus contains proper nouns (politicians, companies, product names) that may date rapidly.

\begin{table}[t]
\centering
\begin{tabular}{lrr}
\toprule
Split      & \# records & Share \\
\midrule
Train      & 22{,}194   & 81.0\% \\
Validation &  2{,}466   &  9.0\% \\
Test       &  2{,}740   & 10.0\% \\
\midrule
\textbf{Total} & \textbf{27{,}400} & 100\% \\
\bottomrule
\end{tabular}
\caption{\textsc{Naver-News-KO} splits.}
\label{tab:splits}
\end{table}

\section{Corpus Statistics}
\label{sec:stats}

\begin{table}[t]
\centering
\small
\begin{tabular}{lrrrr}
\toprule
Field         & min & median & mean & max \\
\midrule
\texttt{title}   length (chars)     &  2 &  29.0 &   29.0 &    84    \\
\texttt{document} length (chars)    & 29 & 855.0 &  999.9 & 16{,}889 \\
\texttt{summary} length (chars)     & 59 & 171.0 &  183.1 &  1{,}838 \\
\bottomrule
\end{tabular}
\caption{Length statistics per text field, computed over all 27{,}400 records via \protect\path{scripts/run_baselines.py} (\texttt{stats} subcommand). Document lengths are long-tailed (mean $\gg$ median); the \texttt{summary} field follows the same pattern.}
\label{tab:lengths}
\end{table}

\begin{figure}[t]
\centering
\includegraphics[width=\linewidth]{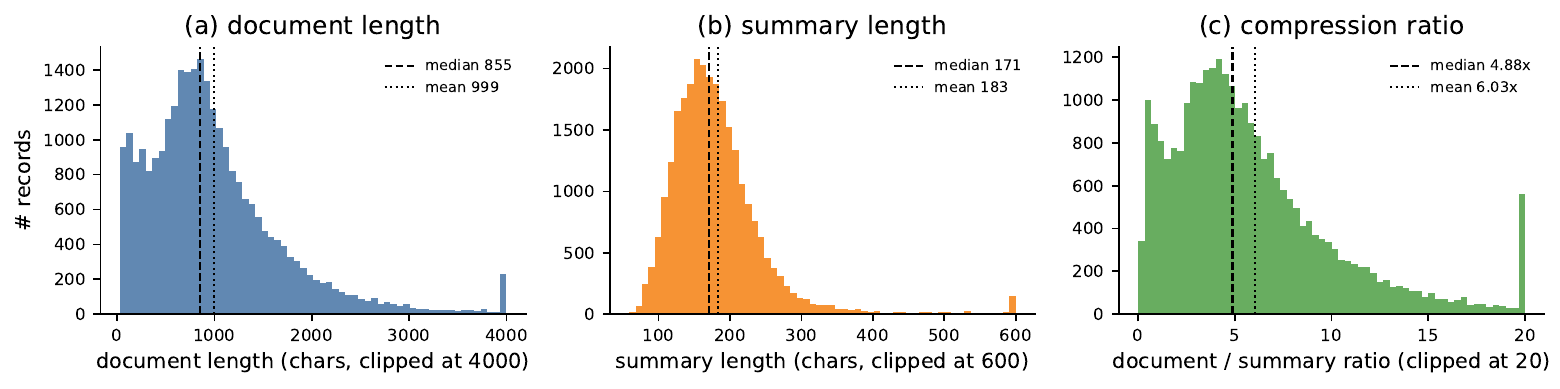}
\caption{Per-record character length distributions for (a)~\texttt{document} and (b)~\texttt{summary}, and (c)~the per-record compression ratio. All panels are aggregated over all 27{,}400 records; dashed lines mark medians, dotted lines mark means. Tails are clipped at 4{,}000 / 600 / 20 respectively for display only; the statistics in Table~\ref{tab:lengths} use the unclipped values.}
\label{fig:length-hist}
\end{figure}

\paragraph{Length.} The \texttt{document} field ranges from 29 to 16{,}889 characters (median 855, mean 1{,}000); the \texttt{summary} field ranges from 59 to 1{,}838 characters (median 171, mean 183). The distributions are long-tailed in both fields (Figure~\ref{fig:length-hist}a,~b). The mean of the per-record character compression ratio $r_i = \lvert\texttt{document}_i\rvert \,/\, \lvert\texttt{summary}_i\rvert$ is \textbf{6.03$\times$} (median 4.88$\times$; Figure~\ref{fig:length-hist}c), comparable in magnitude to English news datasets such as CNN/DailyMail but driven here by longer source documents rather than shorter summaries. Note that the mean of per-record ratios (6.03) differs from the ratio of marginal means (999.9 / 183.1 = 5.46) because the compression ratio is super-linear under the length distribution.

\paragraph{Category balance.} Because only two categories were crawled and their share is uneven (77\% \textit{Economy} vs.\ 23\% \textit{IT/Science}; \S\ref{sec:collection}), \textsc{Naver-News-KO} should not be used to measure \emph{cross-domain} generalization for Korean news summarization. A sharded evaluation by category is appropriate for in-domain work but inherits the 3.3:1 size asymmetry.

\paragraph{Extractive overlap.} News summaries produced by press-editorial processes tend to exhibit high lead-bias; \textsc{Naver-News-KO} is no exception. A Lead-3 extractive baseline (first three sentences of \texttt{document}) achieves ROUGE-1 of 55.1 and ROUGE-L of 50.6 on the test split (Table~\ref{tab:baselines}), a strong floor that abstractive models should be positioned against.

\section{Baselines}
\label{sec:baselines}

We report summarization quality for a small set of models evaluated on the held-out test split ($n=2{,}740$). ROUGE-1 / ROUGE-2 / ROUGE-L~\cite{lin2004rouge} and BERTScore-F1~\cite{zhang2020bertscore} are reported; configuration is fully specified in Table~\ref{tab:baselines}'s caption. All rows are computed from the released scripts (\path{scripts/run_baselines.py}) on a single seed; seed-variance is not characterized here and should be estimated by downstream users when comparing to their own fine-tunes (\S\ref{sec:limitations}).

\begin{table}[t]
\centering
\small
\begin{tabular}{lcccc}
\toprule
Model & R-1 & R-2 & R-L & BERTScore-F1 \\
\midrule
Lead-3 (extractive)                     & 55.12 & 33.22 & 50.60 & 79.72 \\
KoBART~\cite{kobart,lewis2020bart}      & 56.60 & 36.91 & 53.92 & 81.51 \\
Gemma-2B-ko (LoRA SFT)~\cite{gemma2024} & 55.27 & 34.40 & 52.15 & 78.34 \\
\bottomrule
\end{tabular}
\caption{Baselines on \textsc{Naver-News-KO} test split ($n=2{,}740$), greedy decoding, single seed. ROUGE F-scores (\%) are computed with the official \texttt{rouge\_score} implementation~\cite{lin2004rouge}; BERTScore-F1 (\%) uses \protect\path{klue/roberta-base}~\cite{zhang2020bertscore} as the reference encoder with no baseline-rescaling, so the values are comparable within this table only and should \emph{not} be compared directly to English BERTScore values reported with \texttt{roberta-large}. KoBART is \protect\path{gogamza/kobart-base-v2} fine-tuned on our train split for 3 epochs at batch size 8, LR $3\!\times\!10^{-5}$, warmup 0.05; Gemma-2B-ko is \protect\path{beomi/gemma-ko-2b} with LoRA rank 16 on query/key/value/output projections, 3 epochs, batch size 2 with 8-step gradient accumulation, LR $2\!\times\!10^{-4}$, warmup 0.03. Both were trained on a single NVIDIA A100~80GB (KoBART ${\approx}25$~min, Gemma-2B-ko LoRA ${\approx}2.5$~h). We also evaluated \protect\path{gogamza/kobart-summarization} (a KoBART variant already pre-fine-tuned on Korean news summarization), further fine-tuned on our train split; that run produced R-1 45.84 / R-2 23.93 / R-L 44.24 / BS-F1 40.63, a BERTScore that is inconsistent with its ROUGE profile and that we attribute to a decoding/padding artefact under investigation; the row is therefore excluded from the table pending a clean re-run (script and logs are released for inspection).}
\label{tab:baselines}
\end{table}

\paragraph{Interpretation of the Lead-3 floor.} Lead-3---simply emitting the first three sentences of the article body as the summary---already achieves 55.1 ROUGE-1 and 50.6 ROUGE-L, confirming the strong lead-bias common to news summarization corpora derived from editorially produced abstracts (\S\ref{sec:stats}). Abstractive systems on this dataset should be assessed primarily by how far they improve over this extractive floor, not by their absolute ROUGE alone.

\paragraph{KoBART fine-tune.} Three-epoch fine-tuning of KoBART improves Lead-3 by +1.5 R-1, +3.7 R-2, +3.3 R-L, and +1.8 BERTScore-F1. The bigram-level (R-2) gain dominates, suggesting that the abstractive model adds value primarily by recombining content across lead sentences rather than by generating content absent from the lead. This is consistent with the editorial summary provenance discussed in~\S\ref{sec:collection}.

\paragraph{Gemma-2B-ko LoRA SFT.} LoRA supervised fine-tuning of a Korean-tuned Gemma-2B backbone, despite producing visibly more fluent Korean paraphrases than KoBART on sampled outputs, scores lower across all four metrics: --1.3 R-1, --2.5 R-2, --1.8 R-L, and --3.2 BERTScore-F1 relative to KoBART. We attribute the gap to an interaction between (a) the editorial, high-lead-bias nature of the reference summaries, for which near-extractive behavior is rewarded, and (b) the causal-LM fine-tune's tendency to paraphrase the lead rather than copy it. The gap is a property of the corpus more than of the model; a separate human-eval study would be needed to rank these systems on summary quality rather than on reference-overlap metrics.

\paragraph{Purpose of the baseline table.} These numbers are not intended to establish a leaderboard. They exist so that a downstream user fine-tuning a new Korean summarization model can compare against a reported floor from a documented setup, rather than against a blog post.

\section{Usage Evidence}
\label{sec:usage}

As of May~2026, the Hugging Face Hub reports the following for \path{daekeun-ml/naver-news-summarization-ko}: approximately 33{,}000 monthly downloads, 63 likes, and an active community discussion thread. Publicly released community models fine-tuned on this dataset include \path{yunhomaeng/gemma-2b-it-sum-ko-yh} and \path{drlee1/gemma2-9b-it-qdora-summary}. These adoption indicators were the primary motivation for producing this document: downstream users have been citing a dataset that had no citable document.

\section{License and Intended Use}
\label{sec:license}

\paragraph{Dataset artifact.} The packaging (schema, splits, and the statistics manifest produced by this work) is released under \textbf{Apache~2.0}.

\paragraph{Source content.} The \texttt{document} and \texttt{summary} fields contain excerpts of Korean news articles whose underlying copyright is held by the original press outlets, not by the dataset author. The Apache~2.0 license on the artifact does \emph{not} override the copyright status of the article content. Korean copyright law does not recognize US-style fair use and permits only a narrower \textit{공정이용} (fair-dealing) clause (Copyright Act Art.~35-5, 2011). We therefore make no warranty that redistribution of the article text is authorized by the underlying rights-holders, and distribute the dataset for non-commercial research use only, consistent with the prevailing practice of academic NLP corpora derived from openly indexed news aggregators.

\paragraph{Recommended redistribution.} For downstream work wishing to build on \textsc{Naver-News-KO} in a licensing-robust manner, we recommend (i) relying on the \texttt{link} field and a reproducible re-fetch script rather than redistributing the raw \texttt{document}/\texttt{summary} text, or (ii) redistributing only compression statistics and model outputs rather than article text.

\section{Reproducibility}
\label{sec:reproducibility}

The dataset is hosted at
\begin{center}
\url{https://huggingface.co/datasets/daekeun-ml/naver-news-summarization-ko}~\cite{naver_news_ko_card}
\end{center}
A statistics manifest (\path{results/stats.json}: per-field length summaries and compression ratio), a per-category/press/date manifest (\path{results/stats_extra.json}), a leakage-audit manifest (\path{results/leakage.json}), the length-histogram figure (\path{results/fig_length_hist.pdf}), and a baseline training/evaluation script (\path{scripts/run_baselines.py}) are released alongside the source of this document. The script exposes the subcommands: \texttt{stats} (populating Table~\ref{tab:lengths}), \texttt{baseline --model \{lead3,kobart,gemma2b\}} (populating Table~\ref{tab:baselines}), and \texttt{leakage} (populating the near-duplicate audit in \S\ref{sec:collection}). The \texttt{kobart\_summ} and \texttt{mt5} entry points are retained for reproducibility but are not represented in the table; see Table~\ref{tab:baselines} caption and \path{scripts/run_baselines.py} docstrings for status.

\section{Limitations}
\label{sec:limitations}

\begin{enumerate}[leftmargin=*,itemsep=2pt]
\item \textbf{Narrow time window.} Ten days of crawling (2022-07-01 to 2022-07-10) capture the news agenda of a single week, and within that window the daily volume is itself uneven (\S\ref{sec:collection}); vocabulary and topic distribution shift across seasons and across years.
\item \textbf{Two-category scope, unbalanced.} Only \textit{Economy} (77\%) and \textit{IT/Science} (23\%) are represented. Politics, sports, society, and entertainment are absent; generalization to those domains is not supported by this dataset.
\item \textbf{Summary provenance is editorial, not human-abstractive.} The \texttt{summary} field is the article-level abstract surfaced on the source page, written by the outlet's editorial staff, not a post-hoc human-written summary produced specifically as a supervision target.
\item \textbf{Split construction and wire-reprint leakage.} Random splitting does not guarantee \texttt{press}-disjoint or \texttt{date}-disjoint evaluation, and wire-service reprints across outlets produce measurable near-duplicate overlap (title-Jaccard $\geq 0.8$ on 16.8\% of test examples; \S\ref{sec:collection}). Absolute test scores should therefore be read as optimistic.
\item \textbf{Single-seed baselines.} Table~\ref{tab:baselines} reports a single training seed per row; the $+1.5$~R-1 margin of KoBART over Lead-3 has not been demonstrated to exceed seed-variance on this corpus.
\item \textbf{Reference-metric limits.} ROUGE and BERTScore both reward lexical/semantic overlap with a high-lead-bias editorial summary, so abstractive paraphrase (e.g., Gemma-2B-ko) may be penalized relative to near-extractive output irrespective of summary quality.
\item \textbf{Copyright.} Article bodies remain the intellectual property of their respective press outlets (\S\ref{sec:license}).
\item \textbf{No gold abstractiveness label.} Lead-bias cannot be separated from genuine abstractiveness without additional annotation.
\end{enumerate}

\section*{Acknowledgments}
We thank the maintainers of the Hugging Face Hub and the Korean open-source NLP community whose continued use of the dataset motivated this documentation effort.

\bibliographystyle{plain}
\bibliography{references}

\end{document}